\documentclass[runningheads]{llncs}
\usepackage[utf8]{inputenc}
\usepackage{graphicx}
\usepackage{hyperref}
\usepackage{amsmath}

\usepackage{biblatex}
\title{eGAN: Unsupervised approach to class imbalance using transfer learning}

\author{Ademola Okerinde \and Tom Theis \and Nasik Nafi \and William Hsu \and Lior Shamir}

\institute{Kansas State University, 2164 Engineering Hall, Manhattan, KS 43017-6221, USA  \\
\email{\{okerinde,lshamir,bhsu,theis,nnafi\}@ksu.edu}}

\date{}

\begin{document}

\maketitle

\begin{abstract}
Class imbalance is an inherent problem in many machine learning classification tasks. This often leads to learned models that are unusable for any practical purpose. In this study, we explore an unsupervised approach to address class imbalance by leveraging transfer learning from pre-trained image classification models. To this end, an encoder-based Generative Adversarial Network (eGAN) is proposed which modifies the generator of a GAN by introducing an encoder module and adopts the GAN loss function to directly classify the majority and minority class. To the best of our knowledge, this is the first work to tackle this problem using GAN-based loss function rather than augmenting the dataset with synthesized fake images. Our approach eliminates the epistemic uncertainty in the model predictions, as \(P(minority)\) and  \(P(majority)\) need not sum up to 1. The impact of transfer learning and combinations of different pre-trained image classification models at the generator and the discriminator level is also explored. Best result of 0.69 F1-score was obtained on CIFAR-10 classification task with an enforced imbalance ratio of 1:2500. Our implementation code is available at - \\ \url{https://github.com/demolakstate/eGAN_addressing_class_imbalance_with_transfer_learning_on_GAN.git}.



\end{abstract}

\section{INTRODUCTION}
\label{introduction}

A dataset is considered imbalanced when there is a significant, or in some cases, extreme disproportion between the number of samples of the different classes in the dataset. The class or classes with large number of samples are called the majority, while the class with few examples are denoted as the minority. In many cases, the machine learning model is required to correctly classify the minority class while minimizing the misclassification of the majority class. However, the skewness in the data often leads machine learning classification methods to favour the majority class.

Class imbalance problem in computer vision is normally approached either at the data level or algorithm level. Using data augmentation, a class with a small number of samples can be expanded into a class with a much larger number of samples. Earlier data augmentation was achieved simply by transforming images via scaling, cropping, flipping, padding, rotation, brightness, contrast, saturation level etc \cite{zhang2015convolutional}. Now-a-days, synthetic images can also be generated using generative models such as VAE, GAN \cite{kingma2013auto}\cite{goodfellow2014generative}. As a result, a humongous image dataset can be created from the images of the minority class. 

At the algorithm level, the objective function is tweaked to heavily penalize the network for mis-classifying the minority class \cite{sun2007cost} \cite{ling2008cost}. The most popular is cost-sensitive approach. Here, the classifier is modified to incorporate varying penalty for each of considered groups of examples. By assigning a higher cost to the less represented set of samples its importance is boosted during training.

Transfer learning has been known to help improve the performance of machine learning models \cite{wang2018transferring}. By fine-tuning varying number of layers in the pre-trained image classification model, the pre-trained model can serve as a feature extractor, while adding a classifier head for more specific feature learning for the current task.

In this work, we compared the performance of various pre-trained image classification models for the task of unsupervised image classification with varying imbalance ratios. Our architecture, named eGAN, is developed to serve as a basis for this comparison. Using GAN \cite{goodfellow2014generative}, we reparameterise the task of the discriminator as a classifier which outputs a positive score for majority samples and a negative score for the minority ones. We integrate an encoder module to the GAN network that encodes the minority samples into a latent code from which the generator learns. While most GAN-based architectures focus on the output of the generator, in our proposed approach, as we intuitively adapted the vanilla GAN network and the corresponding loss function to directly classify the majority and the minority class, we are more concerned about the performance of the discriminator.


\section{RELATED WORK}

There have been a lot of work in the last few decades to address class imbalance. Earlier approaches include deliberate undersampling of the majority class or oversampling of the minority class by mere copying \cite{drummond2003c4}. However, for image data the earlier approach leads to loss of useful data information while the latter approach causes overfitting \cite{nafi2020addressing}. Data augmentation via rotation, scaling, cropping etc can be considered as a variant of oversampling which copy the same data, however with little modification \cite{zhang2015convolutional}. VAE and GAN enables the generation of completely new data \cite{kingma2013auto}\cite{goodfellow2014generative}. In recent years, GAN-based approaches have gained much popularity than others and a good number of variants of vanilla GAN have been proposed to address class imbalance \cite{radford2015unsupervised} \cite{zhu2017unpaired} \cite{antoniou2017data}. 

In \cite{huang2020towards}, an ensemble method was proposed based on advanced generative adversarial network to generate new samples for the minority class to restore balance. Our opinion is that the computational demand of such approach is enormous, and many low-income countries of the world do not have access to such computation power. Deep Cascading (DC) with a long sequence of decision trees could help to handle unbalanced data \cite{bria2020addressing}. A DC is a sequence of n classifiers where each sample x passes to the next classifier only if the current one classifies it as positive according to a high-sensitivity decision threshold. However, this works well with foreground-background imbalance unlike the classification task. Transfer learning with GAN was used to generate images from limited data in \cite{wang2018transferring}. Their result showed that knowledge from pre-trained networks can ensure faster convergence and significantly improve the quality of generated images. 



\section{METHODOLOGY AND EXPERIMENTAL
DESIGN}
In this section, we discuss our proposed approach and the various testbeds that were used in our experiments.
\subsection{ADDRESSING CLASS IMBALANCE WITH eGAN}
\label{eGAN}


The proposed architecture is based on adaptation of existing Deep Convolutional Generative Adversarial Network (DCGAN)\cite{radford2015unsupervised} by incorporating an encoder module. This module encodes minority samples in latent space needed by the generator G to generate minority samples that are capable of fooling the discriminator D. On the other hand, the discriminator D is fed with data samples drawn from majority distribution and the generated output of the generator G. D and G are simultaneously optimized through the following two-player minimax game with value function V(G,D) in \ref{loss_equation}.


\begin{equation}
\overset{min}{G} \,\, \overset{max}{D} V(D,G) = E_{X_{ma} \sim P_{ma}}[\log D(X_{ma})] + E_{X_{mi} \sim P_{mi}}[\log (1 - D(G(X_{mi})))]
\label{loss_equation}
\end{equation}
where $X_{ma}$ and $X_{mi}$ are majority and minority sample distributions respectively.
    
Over the course of iteration, the discriminator D is optimized to assign a negative score to the minority data distribution and a positive score to the majority data distribution. This enables the discriminator D to act as a classifier. 

\subsubsection{Encoder-Generator module}

Our latent space is composed of 128 units vector.
Rather than feeding the generator with random noise as is typical of most GAN implementation, we added an encoder module that forces the generator to learn from known distribution (minority distribution). The encoder part consists of the pre-trained DenseNet121 followed by global average pooling layer and latent dimension space. The generator part has two transposed convolutional layers. We use LeakyRelu activation function with alpha set to 0.2; batch normalization and Sigmoid function at the final layer.

\subsubsection{Discriminator module}
\begin{figure}[hbt!]
\centering
\includegraphics[scale=0.35]{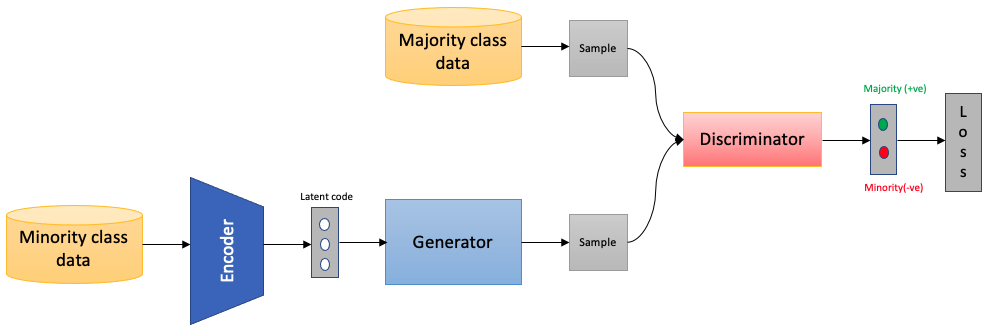}
\caption{Encoder-based Generative Adversarial Network (eGAN) architecture}
\label{fig:encoder-based GAN}
\end{figure}

The pre-trained discriminator has 7,038,529 parameters out of which only 39,937 are trainable. A layer of global average pooling follows the pre-trained DenseNet121. We use a dropout of 0.2 followed by the final one unit dense layer. The overall architecture of our encoder-based generative adversarial network is shown in Figure~\ref{fig:encoder-based GAN}.

\subsection{Selection of pre-trained image classification weights}

We perform experiments on VGG16, VGG19, EfficientNetB2, ResNet101 and DenseNet121 pre-trained classification models on ImageNet dataset. Here, we fine-tuned only top five layers at each of the pre-trained models. Table \ref{tab:experiments_pre_trained_model} shows the maximum precision, recall and F1-score obtained on CIFAR-100 with imabalance ratio 1:50 by using different combinations of pre-trained models.

\begin{table*}[ht]
\begin{center}
\caption{Comparative analysis of different pre-trained models configuration on Generator and Discriminator using CIFAR-100 dataset}
\label{tab:experiments_pre_trained_model}
\begin{tabular}{ | c | c | c | c | c | }
\hline
 Discriminator & Generator & Precision & Recall & F1\\ 
 pre-trained & pre-trained &   &   &  \\ 
 \hline
 ResNet101 & VGG19 & 0.72 & 1.0 & 0.78 \\  
 VGG19 & ResNet101 & 0.73 & 1.0 & 0.69 \\ 
 EfficientNetB2 & VGG19 & 1.0 & 0.22 & 0.32 \\ 
 VGG19 & EfficientNetB2 & 0.7 & 0.86 & 0.71 \\ 
 ResNet101 & VGG16 & 1.0 & 0.17 & 0.27 \\\hline
\end{tabular}
\end{center}
\end{table*}

DenseNet121 \cite{huang2017densely} was used for pretraining our eGAN. After experimenting different pre-trained architectures and different layers of fine-tuning, we obtained best result with fine-tuning only top 5-layer out of 427 layers of DenseNet121.

\subsection{Dataset}
\label{datasets}

Several commonly used datasets were used in this study. In order to model the real-world scenario of heavy imbalance, we used only few samples of the minority class as input to the encoder module. Detail overview is shown in Table \ref{tab:dataset_overview}.

\begin{table*}[ht]
 \begin{center}
\caption{CIFAR-10 and CIFAR-100 dataset overview - CIFAR-100 in parenthesis}
\label{tab:dataset_overview}
\begin{tabular}{ |c |c| c| c| c| }
\hline
 Class imbalance & \# training & \# training & \# testing & \# testing \\ 
  ratio & minor & major & minor & major\\
 \hline
 1:2500 & 2(-) & 5000(-) & 1000(-) & 1000(-)\\ 
 1:1000 & 5(-) & 5000(-) & 1000(-) & 1000(-)\\
 1:500  & 10(1) & 5000(500) & 1000(100) & 1000(100)\\
 1:50   & 100(10) & 5000(500) & 1000(100) & 1000(100)\\ 
 1:1 & 500(500) & 500(500) & 1000(100) & 1000(100)\\
 $^*$1:1 & -(1) & -(1) & -(100) & -(100)\\
 \hline
\end{tabular}
\end{center}
\end{table*}


The CIFAR-10 dataset \cite{cifar10} consists of 60,000 32$\times$32 colour images in 10 classes, with 6000 images per class. There are 50,000 training images and 10,000 test images. Here we use {\it airplane} as minority and {\it automobile} as majority.


CIFAR-100 \cite{cifar10} is similar to CIFAR-10, except it has 100 classes containing 600 images each. Each class has 500 training images and 100 testing images. The 100 classes in the CIFAR-100 are grouped into 20 superclasses. Each image comes with a "fine" label (the class in which it belongs) and a "coarse" label (the superclass).
We also use the pneumonia subset of Stanford CheXpert dataset \cite{irvin2019chexpert} for experimenting on an inherent imbalanced dataset. The dataset contains 4576 and 167407 minority and majority samples, respectively.






\section{RESULTS AND DISCUSSION}
\label{results}

All discriminator scores that are less than zero are classified as minority, otherwise they are classified as majority class. Table~\ref{tab:experiments_pre_trained_model} shows the result obtained by combining various pre-trained models. Adam optimizer with a learning rate of 1e-4 was used to train all models for 100 epochs.

\begin{figure*}[h] 
\centering
  \includegraphics[height=1.8in, width=2in]{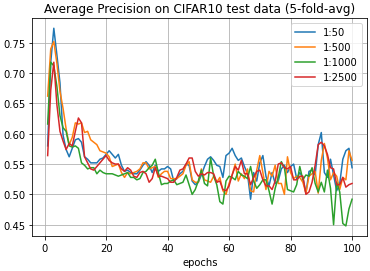}
    \qquad
  \includegraphics[height=1.8in, width=2in]{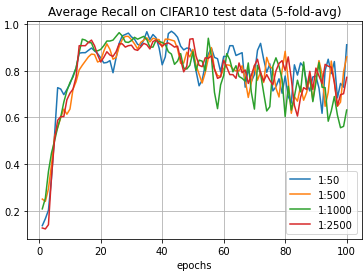}
\vspace{20pt}

  \includegraphics[height=1.8in, width=2in]{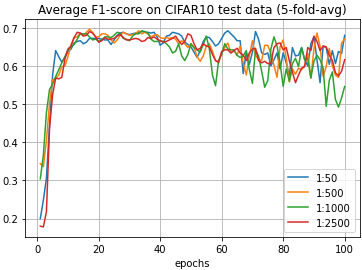}
\qquad
  \includegraphics[height=1.8in, width=2in]{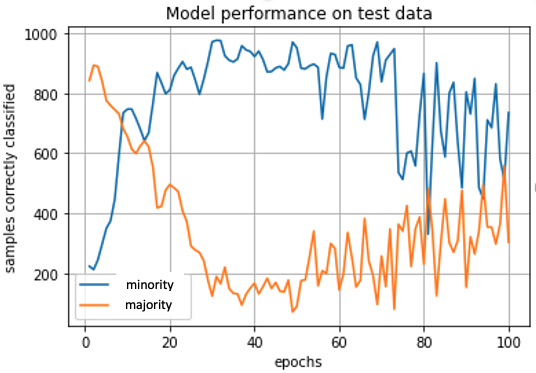}

  \caption{eGAN's average precision (top left), average recall (top right), average F1-score (bottom left), and average performance (bottom right) on CIFAR-10 test dataset using DenseNet121.}
\label{fig:cifar10}
\end{figure*}

As can be seen in Figure~\ref{fig:cifar10}, the model achieved a Nash equilibrium on test data at around 10 epochs. Here, we perform inference on test data at every epoch and plot the number of samples correctly classified. 
Five layers of DenseNet121 were fine-tuned at each generator and discriminator module, while 422 layers' weight were kept fixed. At Nash, the discriminator correctly classified roughly 700/1000 of each of the minority and majority test data. This result convinces us that transfer learning with GAN can be used to overcome the challenge of highly imbalanced dataset, owing to the fact that we train only with 10 samples of the minority class and 5000 samples of the majority class. Similar performance is observed in CIFAR-100 with imbalance ratio 1:50. 


\begin{figure}
\centering
\includegraphics[scale=0.35]{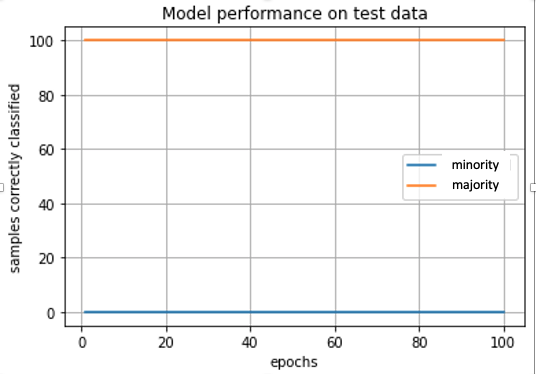}
\caption{Performance of eGAN on \textbf{CIFAR-100} test data using DenseNet121. Only the generator network is pre-trained}
\label{fig:denseNet121_fine_tune_generator_only_cifar100}
\end{figure}

Without pre-training the discriminator, the effect of the high imbalance in the training set is revealed, as the discriminator is skewed towards the majority class in the training set, thereby missing all the minority samples in the test data. This can be seen in Figure~\ref{fig:denseNet121_fine_tune_generator_only_cifar100} on CIFAR-100. This behaviour pattern is observed on CIFAR-10 as well. We experiment with no pre-training at all, neither in the discriminator nor generator, and observed exact same pattern. Therefore, we can safely conclude that the use of transfer learning helps unsupervised image classification in a highly imbalanced domain.

Training can be stopped as soon as Nash equilibrium is reached, as this point gives the model best performance on the minority and majority class. An acceptable threshold can also be set for the absolute difference of the number of correctly classified samples of both classes. For instance, if the $\mid$ correctly\_classified\_minority - correctly\_classified\_majority $\mid$ $\leq$ 20. The precision, recall and F1-score curves on CIFAR-10 averaged over five folds at different imbalance ratios are shown in Figure~\ref{fig:cifar10}.

\begin{figure*}[tb]
\centering
  \includegraphics[height=1.6in, width=1.6in]{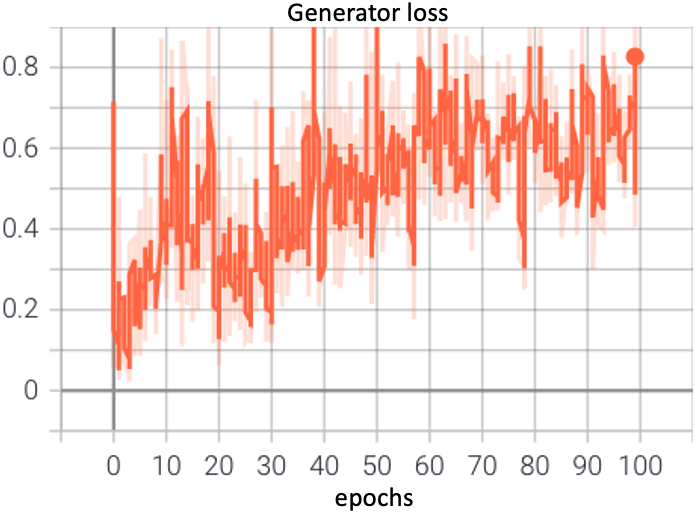}
    \qquad
  \includegraphics[height=1.6in, width=1.6in]{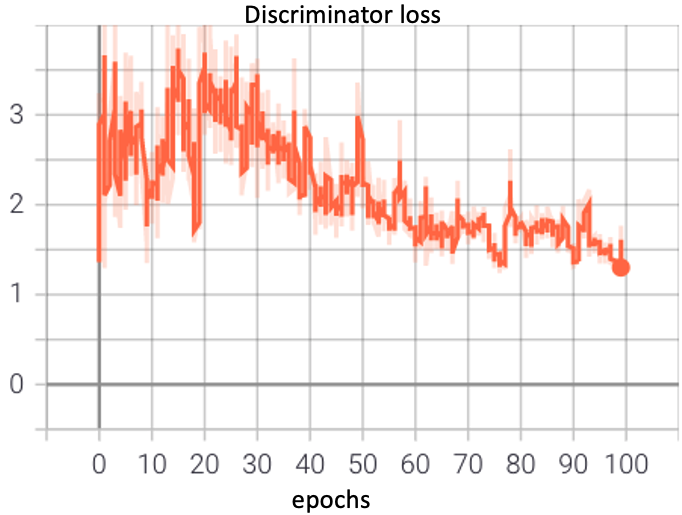}


  \caption{Discriminator and Generator loss.}
\label{fig:disc_gen_loss}
\end{figure*}

We observed that at the early training epochs, typically between 1 and 40 epochs, the generator tries to achieve its objective of fooling the discriminator by generating samples from the majority class fed into the discriminator. That results in more of the minority samples being mis-classified as the discriminator "knows" the distribution of the majority too well. A drastic change occurs when the generator start generating samples from latent vector, which can fool the discriminator as seen in generator and discriminator loss shown in Figure~\ref{fig:disc_gen_loss}.

\subsection{Imbalance ratios}

To eliminate bias in model performance, we conducted 5-fold-cross-validation on the minority samples and average the result. 

\subsubsection{Class ratio of 1:2500}

We experiment on CIFAR-10 dataset by deliberately using an unbalanced subset of the training set. At the 4th epoch, our model correctly identifies 257 minority and 821 majority samples out of 1000 each. At epoch 5, a sharp change occurred that led to 821 minority samples being correctly classified, while only correctly classifying 233 majority samples as shown in Table~\ref{tab:confusion_matrix}. We also observed that at epoch 72 the performance of the network on the majority and minority classes reached a nash equilibrium with a threshold difference of less than or equal to 20.

\subsubsection{Class ratio of 1:1000}

At epoch 81 on CIFAR-10 dataset, nash equilibrium was reached. At this epoch, 532 and 525 minority and majority test data respectively were correctly classified. We observed that the classifier had another major shift between epoch 5 and 6. At epoch 5, best result was obtained. The network was able to classify 863 majority tests and 585 minority tests correctly out of the 1000 samples. At epoch 6, 634 majority and 810 minority tests were classified correctly as swown in Table~\ref{tab:confusion_matrix_1to1000}. Maximum precision, F1-score and recall of 0.88, 0.74 and 0.99 were obtained at epochs 3, 6 and 37, respectively. 

\begin{table}[hbt]
\centering
  \caption{Confusion matrix of imbalance ratio 1:2500}
  \label{tab:confusion_matrix}
  \begin{tabular}{l|c|c}
& Predicted minority & Predicted majority \\
\hline
Actually minority & 821 & 179  \\
Actually majority & 767 & 233  \\
  \hline
  \end{tabular}
\end{table}

\begin{table}[hbt]
\centering
  \caption{Confusion matrix of imbalance ratio 1:1000}
  \label{tab:confusion_matrix_1to1000}
  \begin{tabular}{l|c|c}
& Predicted minority & Predicted majority \\
\hline
Actually minority & 810 & 190  \\
Actually majority & 366 & 634  \\
  \hline
  \end{tabular}
\end{table}

\subsubsection{Class ratio of 1:500}

Both CIFAR-10 and CIFAR-100 were used to experiment imbalance ratio 1:500. On CIFAR-10, maximum precision, recall and F1-score on averaging 5-fold-cross-validation are 0.75, 0.95 and 0.70 respectively as shown in Table \ref{tab:imbalance ratio}. The maximum precision is slightly lower on CIFAR-100 with 0.60. However, the recall and F1-score which are 0.96 and 0.68 are roughly the same.

\subsubsection{Class ratio of 1:50}

We demonstrate our model performance on imbalance ratio 1:50 using CIFAR-100 and CIFAR-10. For CIFAR-100, a sudden change occurred between epoch 69 and 70 as follows majority: 57, minority: 59; and majority: 55, minority: 59. A nash equilibrium is attained at epoch 68, with 56 correctly classified minority as well as majority class. At epoch 97 maximum F1-score and recall of 0.66 and 0.77 were obtained respectively, while maximum precision of 0.6 was obtained at epoch 74.

\begin{table*}[ht]
\begin{center}
\caption{Maximum precision, recall and F1-score on CIFAR-10 and CIFAR-100 (avg. 5-fold) - CIFAR-100 in parenthesis}
\label{tab:imbalance ratio}
\begin{tabular}{ |c| c| c| c|}
\hline
 ratio & Precision & Recall & F1 \\ 
 \hline
 1:2500 & 0.72(-) & 0.94(-) & 0.69(-) \\ 
 1:1000 & 0.72(-) & 0.96(-) & 0.69(-) \\
 1:500  & 0.75(0.60) & 0.95(0.96) & 0.70(0.68) \\
 1:50   & 0.78(0.7) & 0.97(0.86) & 0.69(0.71) \\ 
 1:1 & \textbf{0.82}(0.72) & 0.98(\textbf{1.0}) & 0.72(\textbf{0.78}) \\
 $^*$1:1 & -(0.53) & -(0.86) & -(0.63) \\
 \hline
\end{tabular}
\end{center}
\end{table*}

\begin{table}[ht]
\begin{center}
\caption{Precision, recall and F1-score on pneumonia subset of CheXpert dataset}
\label{tab:chexpert-result}
\begin{tabular}{ |c|c|c|c| c| c| c|c|c|}
\hline
\multicolumn{1}{|c|}{} & \multicolumn{2}{|c|}{\#training} & \multicolumn{2}{|c|}{\#testing} &  \multicolumn{4}{|c|}{} \\
 Model & minor & major & minor & major & imbalance ratio & Precision & Recall & F1 \\ 
 \hline
 eGAN & 30 & 1080 & 1000 & 1000 & 1:36   & 0.51 & 0.97 & 0.67 \\
  baseline & 30 & 1080 & 1000 & 1000 & 1:36   & 0.5 & 1.0 & 0.67 \\
 \hline
\end{tabular}
\end{center}
\end{table}

\subsubsection{Class ratio of 1:1}

We use CIFAR-100 to demonstrate the performance of eGAN on a balanced dataset. We notice that the experimental performance follow the same pattern as imbalanced dataset. Training starts with mostly all the majority correctly classified and all the minority mis-classified. At epoch 48, a nash equilibrium (with threshold less than or equal to 5) is achieved, with 76 and 71 of minority and majority correctly classified respectively. The maximum F1-score of 0.78 is reached at epoch 53 as shown in Table \ref{tab:imbalance ratio}. Instead of using 500 samples each of minority and majority class, by training on a single instance of minority and majority sample ($^*$1:1) of CIFAR-100, we obtained an F1-score of 0.63. This demonstrates the impact of transfer learning on the training.

\subsubsection{Class ratio of 1:36} For pneumonia subset of CheXpert dataset with imbalance ratio 1:36, the best performed model achieves 0.51, 0.97, and 0.67 precision, recall, and F1-score respectively. The results shown in Table \ref{tab:chexpert-result} is evaluated on 1000 of each minority and majority test set. As can be seen in the table, our approach did not beat the baseline classification model because this task is more of an anomaly detection task rather than a classification problem. Also, the pre-trained image classification model source dataset (i.e. ImageNet) is different from the medical domain. Exploring more variants of complex GAN architectures like BigGAN, StyleGAN and ProGAN could possibly help.

\section{CONCLUSION}
\label{conclusion}
In this work, we demonstrates the capability of a GAN-based unsupervised technique to address class imbalance using pre-trained models. We conducts experiment with varying levels of imbalance ratios in the training dataset. Instead of synthesizing artificial images with the generator for data augmentation, we employ the discriminator as a classifier and formulate the loss function accordingly. Experimental results reveal that transfer learning plays a significant role in the model performance. 
The performance measure of interest plays a significant role in deciding the trained model from which epoch to deploy in production as the model at different epochs favour different evaluation metrics for example sensitivity. 
Future work will focus on the usage of this approach for anomaly detection task where the distinguishing features between normal (majority) and abnormal (minority) are less profound. Our work can be further explored in object detection tasks in case of imbalance between foreground and background.


\printbibliography

\end{document}